\documentclass[letterpaper, 10pt, conference]{IEEEtran}  % Comment this line out if you need a4paper
\usepackage{times}

\IEEEoverridecommandlockouts                              % This command is only needed if 
\let \labelindent\relax

\usepackage{newclude}
\usepackage{amsmath}               
\usepackage{amssymb}
\usepackage{adjustbox}
\usepackage{epsfig, psfrag}
\usepackage{amsthm}
\usepackage{latexsym}
\usepackage[
singlelinecheck=false % left-justify captions
]{caption}
\usepackage{bbm}
\usepackage{soul}
\usepackage{mathrsfs}
\usepackage{algorithm}
\usepackage{multicol}
\usepackage[noend]{algpseudocode}
\usepackage{algorithmicx}
\usepackage{amsfonts} 
\usepackage{mathtools}
\usepackage[affil-it]{authblk}
\usepackage{dsfont} 
\usepackage{euscript} 
\usepackage{amsthm}
\usepackage{enumitem}

\algnewcommand\algorithmicinput{\textbf{INPUT:}}
\algnewcommand\INPUT{\item[\algorithmicinput]}

\allowdisplaybreaks
\usepackage{mathtools}
\usepackage{epstopdf}
\usepackage{subfigure}
\usepackage{graphics,color}
\usepackage{graphicx}
\usepackage[table, svgnames, dvipsnames]{xcolor}
\usepackage{lipsum}
\usepackage{soul}
\usepackage[noadjust]{cite}
\usepackage{multirow}
\usepackage{pifont}% http://ctan.org/pkg/pifont
\newcommand{\cmark}{\ding{51}}%
\newcommand{\xmark}{\ding{55}}%

\usepackage{etoolbox}
\makeatletter
\patchcmd{\@makecaption}
  {\scshape}
  {}
  {}
  {}
\makeatletter
\patchcmd{\@makecaption}
  {\\}
  {.\ }
  {}
  {}
\makeatother
\def\tablename{Table}

\definecolor{my_dark_blue}{RGB}{43, 89, 162}
\definecolor{my_orange}{RGB}{255, 153, 0}
\definecolor{my_teal}{RGB}{0, 175, 181}
\definecolor{my_maroon}{RGB}{117, 13, 55}

\makeatletter
\let\NAT@parse\undefined
\makeatother
\theoremstyle{plain}

\begin{document}
\bstctlcite{IEEEexample:BSTcontrol}

\title{NashFormer: Leveraging Local Nash Equilibria for Semantically Diverse Trajectory Prediction }
% Old Title: GAME-UP:  Game-Aware Mode Enumeration and Understanding for Trajectory Prediction
\author{Justin Lidard$^{1,3}$, \thanks{$^{1}$Department of Mechanical and Aerospace Engineering, Princeton University, Princeton, NJ 08540, USA, \texttt{jlidard@princeton.edu}} Oswin So$^{2}$, Yanxia Zhang$^{3}$, Jonathan DeCastro$^{3}$, Xiongyi Cui$^{3}$, Xin Huang$^{2}$\thanks{$^{2}$Massachusetts Institute of Technology, Cambridge, MA 02139, USA} \\ \textnormal{ Yen-Ling Kuo$^{2}$, John Leonard$^{3}$, Avinash Balachandran$^{3}$, Naomi Ehrich Leonard$^{1}$ and Guy Rosman}$^{3}$\thanks{$^{3}$Toyota Research Institute, Cambridge, MA 02139, USA} \thanks{This research has been supported in part by an NSF Graduate Research Fellowship. This article solely reflects the opinions and conclusions of its authors and not TRI or any other Toyota entity.} }
\maketitle

% Page numbers
\thispagestyle{plain}
\pagestyle{plain}

%\date{March 2022}
% \author{
%     Justin Lidard,  
% }

\begin{abstract}
    Interactions between road agents present a significant challenge in trajectory prediction, especially in cases involving multiple agents. Because existing diversity-aware predictors do not account for the interactive nature of multi-agent predictions, they may miss these important interaction outcomes. In this paper, we propose NashFormer, a framework for trajectory prediction that leverages game-theoretic inverse reinforcement learning to improve coverage of multi-modal predictions. We use a training-time game-theoretic analysis as an auxiliary loss resulting in improved coverage and accuracy without presuming a taxonomy of actions for the agents. We demonstrate our approach on the interactive split of the Waymo Open Motion Dataset, including four subsets involving scenarios with high interaction complexity. Experiment results show that our predictor produces accurate predictions while covering $33\%$ more potential interactions versus a baseline model.    
\end{abstract}

\section{Introduction}

Motion forecasting is a critical task in autonomous driving for understanding distinct interactive behaviors of road agents and making safe decisions. In \textit{joint} motion prediction, which involves predicting trajectories for multiple agents simultaneously, the problem of covering many possible outcomes is exacerbated by the existence of many feasible modes of interaction, especially in dense or urban environments. 
Many recent works employ inference-time sampling techniques \cite{gonzalez1985clustering, zhao2021tnt, shi2022motion, luo2022jfp, shi2023mtr++} %
%
% such as farthest point sampling (FPS)  or non-maximum suppression (NMS) \cite{} to
%
to produce a set of representative samples from a much larger proposal distribution, from which diversity can be evaluated post-hoc.

% Motion forecasting involves predicting a future trajectory for a single interested road agent (\textit{marginal} prediction) or multiple road agents simultaneously (\textit{joint} prediction). In recent years, joint prediction \cite{luo2022jfp, shi2022motion, Varadarajan2021MultipathPlusPlus, Ngiam2021SceneTransformer} has received increased attention as it is crucial for safe decision-making in dense or urban environments.

% Covering distinct outcomes with a limited number of samples is a significant challenge in prediction. 

%
% To increase coverage of distinct outcomes, prediction frameworks employ several different strategies. Metrically diverse sampling methods such as farthest point sampling (FPS) \cite{gonzalez1985clustering} adopt a sampling procedure to optimize some pairwise metric between candidate trajectories, typically the $l_2$-norm, while non-maximum suppression (NMS) \cite{zhao2021tnt, shi2022motion, luo2022jfp, shi2023mtr++}, which has received increased attention during recent years, selects the highest-weighted trajectory within a given neighborhood. While metric diversity increases the number of seemingly distinct motion modes, predictive accuracy may suffer as sampled trajectories may not have high likelihood. 
%
Other recent works promote sample diversity instead by affording exploration in the space of discrete semantic representations, typically waypoints \cite{zhao2021tnt, gu2021densetnt} or maneuvers \cite{huang2020diversitygan, huang2022hyper}. However, an explicit latent representation could limit predictive diversity: predicted waypoints may be highly dependent on the type of interaction mode, and predicted maneuvers are limited to %hand-labeling 
a \textit{taxonomy} of hand-labeled behaviors, which may be difficult and time-consuming to produce.

\newlength{\tempdima}
\newcommand{\rowname}[1]% #1 = text
{\rotatebox{90}{\makebox[\tempdima][c]{\textbf{#1}}}}

\begin{figure}[t!]
\centering
\settoheight{\tempdima}{\includegraphics[width=.22\textwidth]{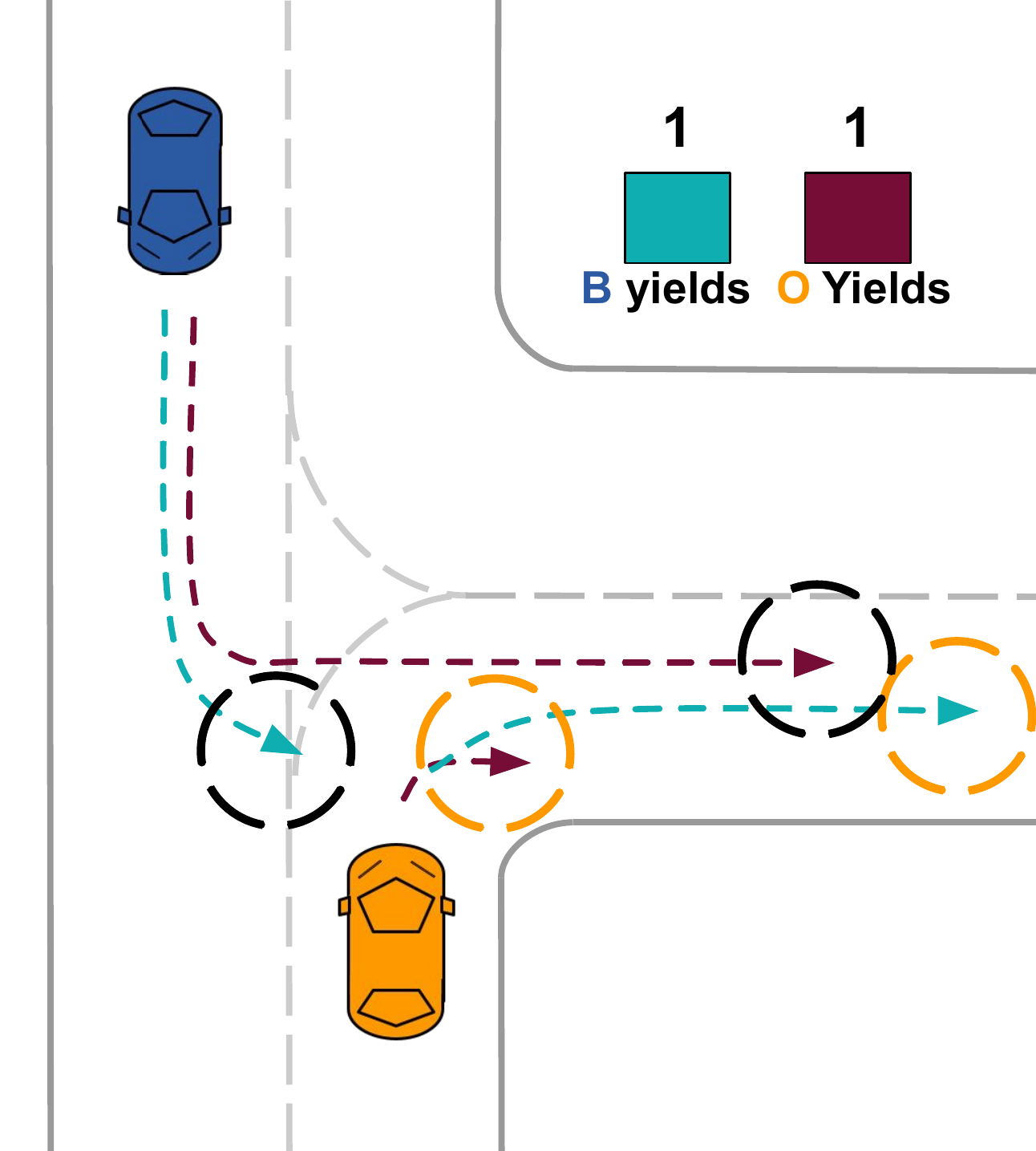}}
\begin{tabular}{@{}c@{ }c@{ }c}
\rowname{Rollout} & \subfigure[]{\includegraphics[width=0.22\textwidth]{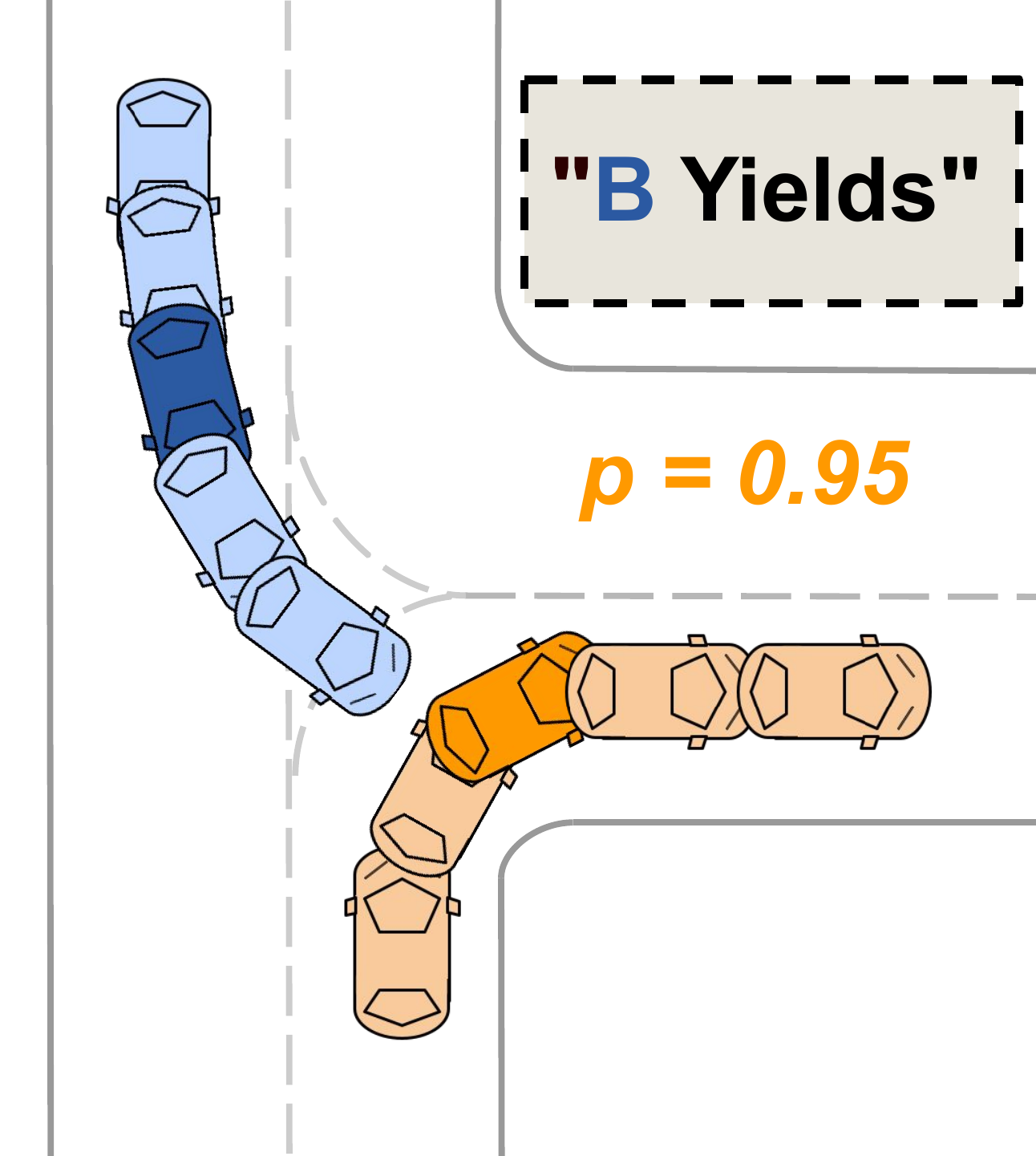}} &
\subfigure[]{\includegraphics[width=0.22\textwidth]{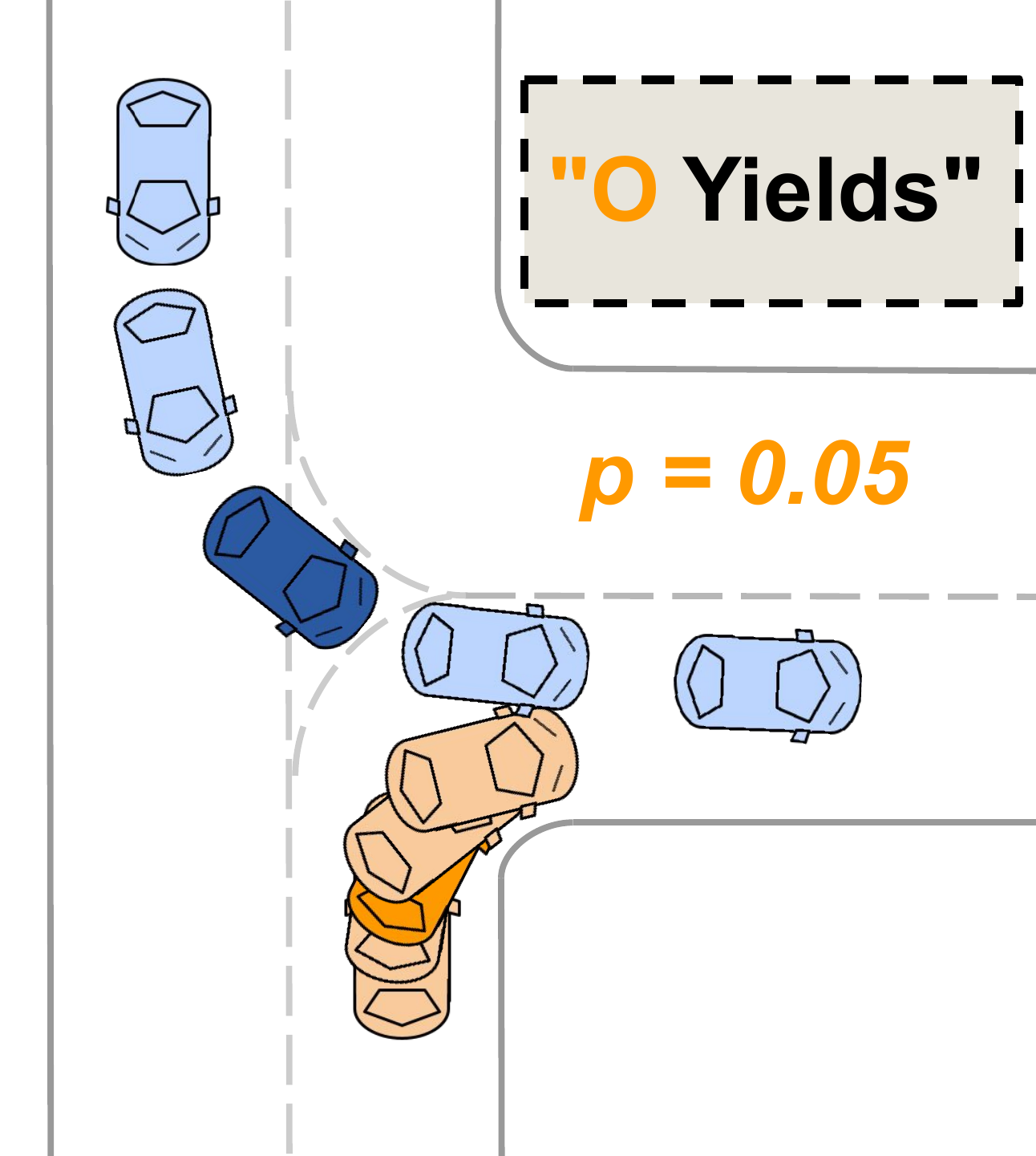}}  \\
\rowname{Prediction} & \subfigure[]{\includegraphics[width=0.22\textwidth]{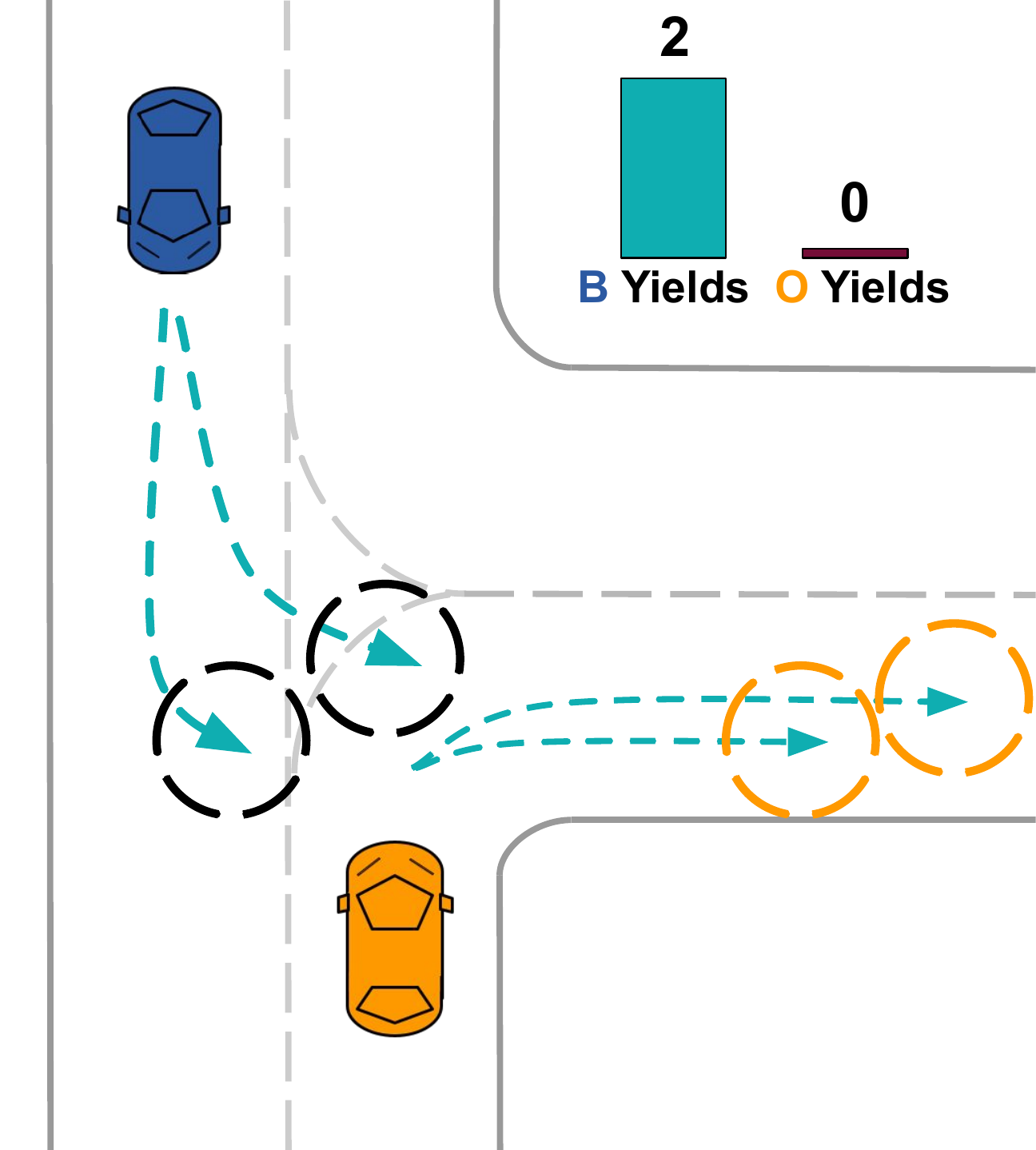}} &
\subfigure[]{\includegraphics[width=0.22\textwidth]{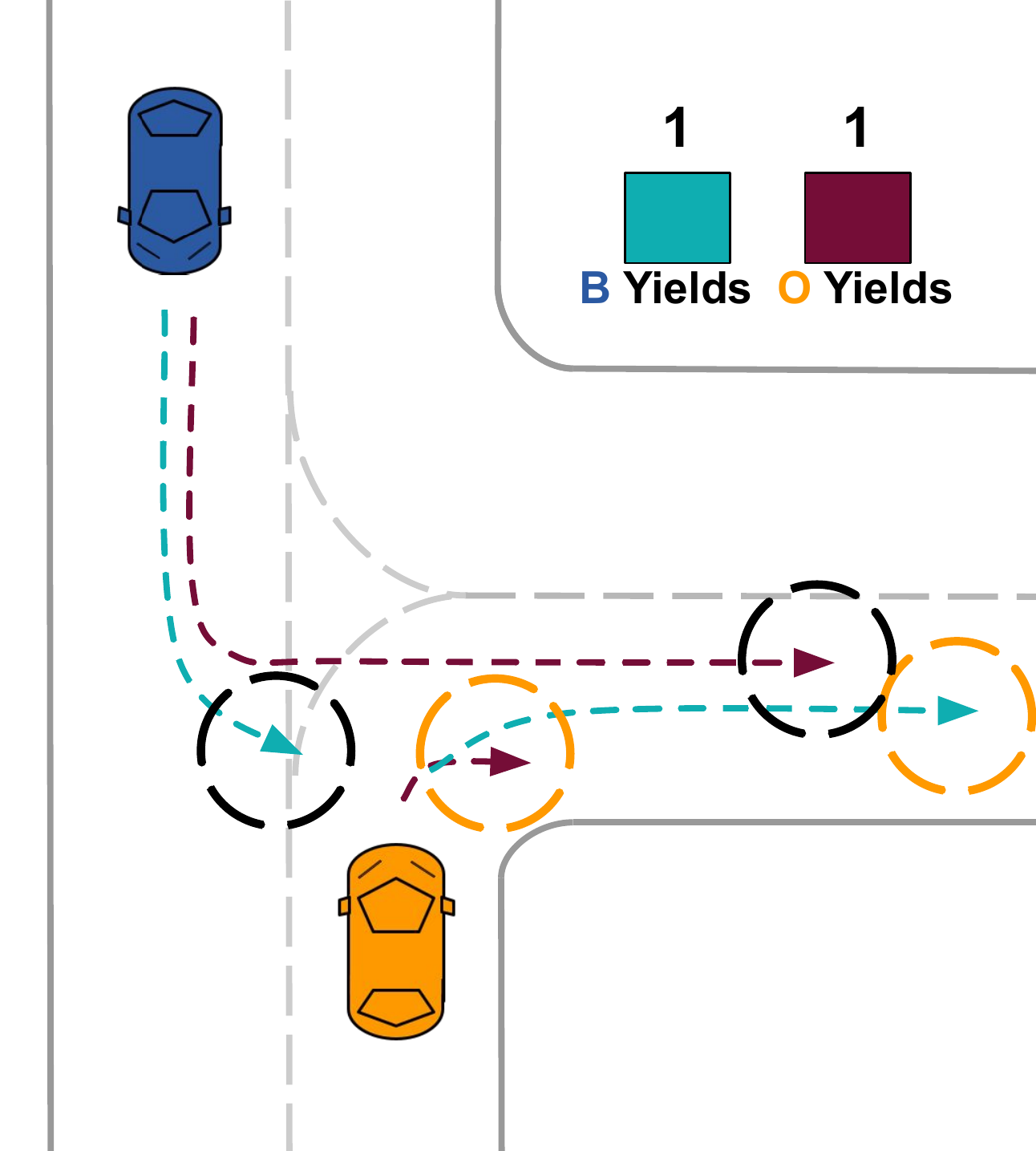}}
\end{tabular}
\caption{ An illustrative example where agents \textcolor{orange}{O} and \textcolor{my_dark_blue}{B} navigate an intersection with different semantic interaction modes: (a) with higher probability ($p=0.95$), \textcolor{my_dark_blue}{B} yields to \textcolor{orange}{O} with both agents respecting right-of-way, and (b) with lower probability ($p=0.05$), \textcolor{orange}{O} yields to \textcolor{my_dark_blue}{B} after \textcolor{my_dark_blue}{B} ignores oncoming \textcolor{orange}{O}, inconveniencing \textcolor{orange}{O}. We show two prediction strategies with limited samples ($K=2$):  (c) a generic predictor uses a metric sampling method such as non-maximum suppression \cite{zhao2021tnt} (dashed circles) and produces two likely but similar trajectories (\textcolor{my_teal}{both teal}), and (d) our method captures \textit{payoff} diversity and finds both distinct outcomes (\textcolor{my_teal}{teal}, \textcolor{my_maroon}{maroon}) \textit{without predicting the semantic label}. \vspace{-0.3in}}
\label{fig:teaser}
\end{figure}
%This affords exploration in the space of the latent representation, but such representations must be hand-labeled or learned from data. 
%Moreover, these latent representations are static and may be difficult to classify in the presence of dynamic interactions, where waypoints and maneuvers are dependent on the actions of other agents.  
There remain several open challenges in multi-modal prediction methods in learning meaningful semantic behaviors without a comprehensive taxonomy or cumbersome hand-labeling, and in creating sampling techniques where sample diversity is evaluated and optimized at inference time. Our insight is to use game-theoretic techniques both for evaluating semantically distinct behaviors and for sampling optimally diverse trajectories, without explicitly labeling distinct outcomes with a manually defined taxonomy. 

Fig.~\ref{fig:teaser} illustrates a simple motivating example wherein existing predictors may fail to capture interactions of interest (failing to yield, Fig.~\ref{fig:teaser}b).
% This is because these approaches only account for high probability events, i.e., the most likely (global, or near-global) Nash equilibria.
The distinct outcomes of such interactions, while intuitive for human drivers, are not captured via the generic predictor (Fig.~\ref{fig:teaser}c), which uses a sampling method such as non-maximum suppression (NMS) to find metrically distinct trajectories, but falls short of finding the other semantic mode where the orange agent yields. A game-aware predictor such as NashFormer will capture both semantic scenarios.

% Game-theoretic approaches have recently proved valuable, as evidenced in simulating human behaviors for prediction \cite{ma2017forecasting, wang2021game}, autonomous vehicle planning under stochastic policies \cite{schwarting2019social,schwarting2021stochastic,so2022multimodal}, and making robots respect social norms in the presence of humans \cite{tian2022safety, wang2022co ,galati2022game}. A key assumption from noncooperative game theory is that agents' actions are constrained by other agents: often, a given agent cannot find a globally cost-minimizing trajectory in the presence of other self-interested agents. Instead, agents operate at \textit{Local Nash Equilibrium (LNE)}, and there may be several distinct LNE in a scenario. We solve for LNE analytically while still permitting stochastic decisions under the standard \textit{bounded-rationality} assumption \cite{Simon1957-md, dequech2001bounded}. 

In this paper, we develop and implement NashFormer: a methodology that leverages the fundamental idea that classification of candidate trajectory predictions according to nearby local Nash equilibria (LNEs) promotes \textit{tunable} diversity at training time, and permits \textit{inference-time} diverse sampling. 
% enumerating and ranking semantically diverse Nash equilibria can significantly improve the performance of multi-agent prediction. 
We design a method for covering equilibria that are diverse in terms of learned utility and show that our method provides \textit{semantic} diversity, without requiring hand-labeled semantic features. 

% Our approach learns a general agent-level utility model and uses local optimization of rollouts to find and score LNE. The predictor is trained to cover the scored LNE using a novel loss. 
Our main contributions are as follows: 
\begin{itemize}
    \item A prediction model that covers semantically distinct LNE,
    \item A game-theoretic analysis module for evaluating, scoring and clustering candidate candidate trajectories,
    \item Experimental results on the Waymo interactive dataset, including strongly interactive subsets of the dataset. These demonstrate the accuracy and coverage of our predictions and provide insight into the influence of model parameters. 
\end{itemize}

\begin{figure*}[t!]
    \centering
    \includegraphics[width=\textwidth]{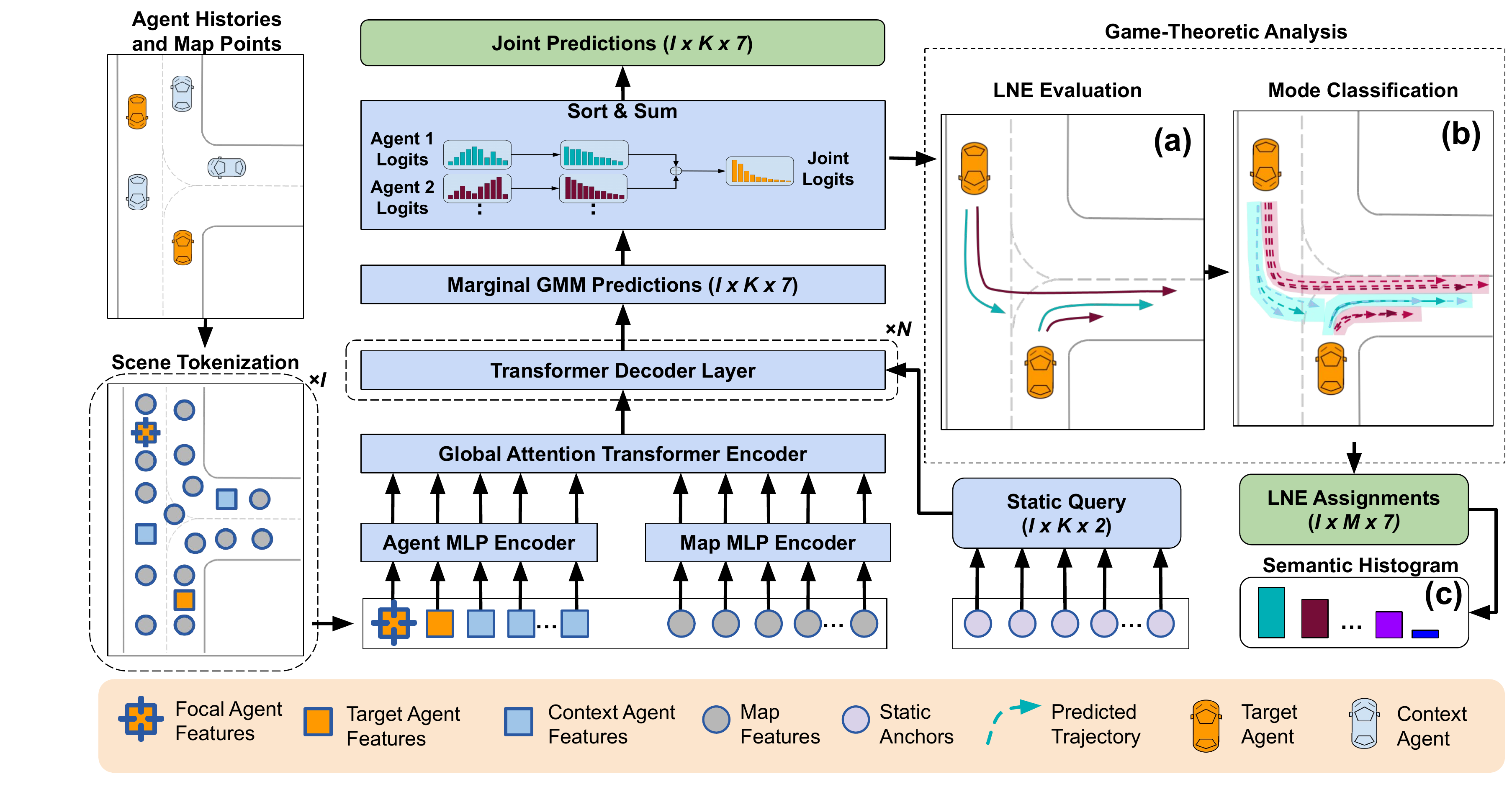}
    \caption{Prediction system training architecture. Multi-agent trajectories and map polylines are tokenized in the local coordinates of \textit{each} focal agent (\textcolor{my_dark_blue}{blue outline}) in the target agent set (\textcolor{my_orange}{orange}) and processed in parallel using a network with shared weights. The encoded features are cross-attended using a transformer encoder. $N$ transformer decoder layers refine the $K$ candidate modes and produce a Gaussian mixture model motion forecast (position mean, position variance, mode weight, and velocity). A game-theoretic analysis scores the semantically meaningful modes: (a) candidate trajectories are optimized for their utility to find LNE, (b) modes are clustered according to the nearest LNE, and (c) the assigned modes are summarized in a semantic histogram. The final transformer decoder layer produces the final prediction for post-processing.  \vspace{-0.2in}}
    \label{fig:predictor_training}
\end{figure*}

\section{Related Work}

Our work relates to several topics in game-theoretic planning and joint trajectory prediction.

\subsection{Dynamic Games and Inverse Optimal Control}

Dynamic games \cite{bacsar1998dynamic} are a common framework for analyzing $N$-player driving scenarios \cite{Fisac2019-hierarchical, Fridovich-Keil2020efficient,wang2021game,so2022multimodal}. Optimal control policies satisfy \textit{Local Nash Equilibrium} (LNE), where each player's expected payoff is locally optimal with respect to their control strategy. %, contrasting with global Nash Equilibria, which are generally computationally intractable \cite{Facchinei2010generalizednash}. 
LNEs have been studied extensively in autonomous driving as a model for interaction in merging \cite{geiger2021learning,cleac2020algames}, highway overtaking \cite{le2021lucidgames,Fisac2019-hierarchical}, and racing \cite{schwarting2021stochastic,wang2021game}. Recent works in game-theoretic planning give efficient methods for solving LNE for deterministic policies \cite{cleac2020algames} and stochastic policies (i.e. \textit{bounded rationality}) \cite{mehr2021maximum}. Extensions have also been presented to games where agents' actions are determined under partial state observations \cite{schwarting2021stochastic} or  by some latent parameters in the policy \cite{schwarting2019social, espinoza2022deep} or cost \cite{le2021lucidgames}. Some recent works \cite{so2022multimodal, so2023mpogames} have addressed \textit{multi-modal} planning under a discrete latent variable, but providing efficient and context-dependent methods for trajectory prediction under many possible modes of interaction is an open challenge. 

Inverse optimal control and reward learning have been  used, even at the single agent level, for understanding human decision-making \cite{ziebart2008maximum, wulfmeier2017large,phan2022driving, evens2021}, facilitating analysis of \textit{bounded rational} decision making for road agents under a maximum entropy (MaxEnt), non-greedy framework. 
The reward-learning approach most similar to ours in spirit is that of \cite{geiger2021learning}. In \cite{geiger2021learning}, the parameters of a quadratic potential game are learned, and then the optimal policies of each agent are solved online to predict the outcome of a highway merge scenario between two agents. Our work is different in three ways: (i) we permit stochastic (bounded-rational) learned policies under the maximum-entropy framework, (ii) we leverage inverse reinforcement learning (IRL) to learn LNE during training and efficient evaluate coverage during inference time (iii) we evaluate our framework on real driving scenarios of varying complexity.

%  We introduce three novel modules: (a) the Joint Policy Evaluation module clusters each set of multi-agent trajectories as a distinct motion mode; (b) the Local Mode Optimization finds the highest-scored trajectory within each cluster, and (c) the Mode Histogram summarizes the candidate modes by their learned Boltzmann likelihood and supervises the final trajectory weights.

% \vspace{-0.1in}
\subsection{Multi-agent Prediction}

%In \cite{Nayakanti2022Wayformer}, the authors design a family of models using self-attention to fuse modalities across temporal and spatial dimensions and cross-attention decoder to produce a diverse set of trajectories; they then study three variations of the scene encoder that differ in how and when different input modalities are fused.  

Within the field of trajectory prediction, 
agent-agent interactions are a critical consideration when scaling to the multi-agent setting \cite{helbing1995social}.
% Modern joint predictors making use of attention-based architectures for fusing multimodal scene, map, agent, and interaction information \cite{Ngiam2021SceneTransformer}. 
The output of a joint predictor may be a raw trajectory sample \cite{gu2021densetnt}, a weighted set \cite{Ngiam2021SceneTransformer}, or a mixture model over discrete modes \cite{Varadarajan2021MultipathPlusPlus}.
Finally, some approaches explicitly model discrete agent interactions in order to better account for them and improve accuracy
\cite{kumar2020interaction, sun2022m2i,ban2022deep}.
% A game-theoretic prediction work is that of \cite{ma2017forecasting}, which leverages an MDP policy model and fictitious play for posterior distribution for (unimodal) pedestrian prediction.
Our work, however, specifically emphasizes \textit{multi-modality} and offline rollouts to replace fictitious play in the prediction framework. 

\subsection{Diverse Prediction}

In sample-based prediction, maintaining diversity of samples to represent distinct outcomes is an active area of research. Farthest Point Sampling (FPS) \cite{huang2020diversitygan,shiroshita2020behaviorally}, Non-Maximum Suppression (NMS) \cite{zhao2021tnt}, and/or neural adaptive sampling \cite{huang2022hyper} provide methods for encouraging outcome diversity using pairwise distances between trajectories as a metric.
Diversity in the sampling is often aided by specific underlying representation, such as a latent layer in \cite{salzmann2020trajectron++}, a mixture model \cite{huang2019uncertainty}, a set of anchor points \cite{shi2022motion}, or the use of a bagging algorithm on the trajectories \cite{phan2020covernet}. 
Several works~\cite{huang2022tip, capo} show that knowledge of downstream tasks allows adaptation of the prediction samples so as to improve results.
In our paper we show that with learned game-theoretic utilities, our prediction framework achieves better coverage of semantic interactions without the need for explicit taxonomies or task definition for semantic coverage.

\include*{subtex/problem_formulation}

\section{Approach}

This section details our procedure for approximating the LNE of the system by modeling it as an inverse maximum-entropy dynamic game. As shown in Fig.~\ref{fig:predictor_training}, we use the approximated LNE to classify candidate trajectories according to their nearest LNE and compute a (differentiable) empirical histogram that represents the relative coverage of each LNE across the full sample set. The empirical histogram is compared using the Kullback-Liebler (KL) divergence to a tunable ``ideal'' histogram determined by the game-theoretic likelihood of the LNE. The KL-divergence score functions as an auxiliary coverage loss during training. At inference time, the empirical histogram may be used to mask duplicate interactive behaviors in a proceedure similar to non-maximum supression (NMS). 
% NashFormer leverages a transformer  \cite{vaswani2017attention}  encoder-decoder model, as depicted in Fig.~\ref{fig:predictor_training}. Like other prediction architectures, NashFormer relies on both agent trajectory and map inputs. Section \ref{approach: GT} discusses preliminaries for inverse game-theoretic reinforcement learning. Section \ref{appraoch: Mode Opt} discusses the process for evaluating and clustering LNE, corresponding to Fig.~\ref{fig:predictor_training}(a-c). Section \ref{Appraoch: model details} discusses the architecture details for NashFormer.
% Section \ref{Approach: Loss Definitions} defines the losses used by NashFormer. 

\subsection{Inverse Maximum-Entropy Dynamic Games}

\label{approach: GT}

In order to characterize the LNEs in each scenario, we first approximate road agents' behavior as a discrete dynamic game. The joint state of the dynamic game follows a nonlinear dynamics equation:
\begin{equation} \label{eqn: dynamics}
    \mathbf x_{t+1} = f_t(\mathbf x_t, \mathbf u_t) \quad \forall t
\end{equation}
where $\mathbf x_t = (x_t^1, ..., x_t^{I})$ denotes the joint state of the system, $\mathbf u_t = (u_t^1, ..., u_t^{I})$ denotes the control applied by all agents, and $I$ is the number of decision-making agents. Under Eqn.~\eqref{eqn: dynamics}, agents affect the joint system simultaneously and non-cooperatively, and we model equilibria as Nash.

% % Note that the Nash 

% over other notions of equilibria, e.g. Stackelberg \cite{Fisac2019-hierarchical}. 

% Under the deterministic and controllable assumption, we use the shorthand $u_{t}^i := x_{t+1}^i$, since state transitions are known \textit{a priori} from a fixed dataset. 

NashFormer learns the agent-level best-response model that evaluates the utility of an action $u_t^i$ by agent $i$ conditioned on the joint state $\mathbf x_t$, contrasting other direct-solving methods \cite{cleac2020algames, Fridovich-Keil2020efficient} %, which may be slow to run at inference time, 
which may have poor coverage of equilibria. The model computes the log-likelihood actions for each agent $i$, based on a  maximum entropy objective:
\begin{equation} \label{eq:payoffs}
    J^i(\pi^i) = \mathbb E^{\pi} \Bigg[ \sum_{t=1}^T r^i(\mathbf x_t, \mathbf u_t) + H[\pi^i(\cdot | \mathbf x_t) ]\Bigg],
\end{equation}
where $\sum_{t=1}^T r^i(\mathbf x_t, \mathbf u_t)$ is the cumulative reward attained by agent $i$ over the joint trajectory $\boldsymbol{\tau}$ of length $T$, and $H[\pi^i(\cdot | \mathbf x_t) ]$ is the Shannon entropy of agent $i$'s policy. We learn the policy $\pi^i$ for each agent $i$ that maximizes the objective in Eqn.~\eqref{eq:payoffs}.
% and assume agents follow Local Nash Equilibrium (LNE) \cite{bacsar1998dynamic}.

% To solve eqn.~\eqref{eq:payoffs}, we assume agents follow Local Nash Equilibrium (LNE) \cite{}, which is defined as a joint policy $\pi$ where agent payoffs are locally optimal when other agents' policies are held fixed, i.e.
% \begin{equation} \label{eqn:LNE}
%     J^i(\pi^i, \pi^{\neg i}) \geq J^i(\pi^{*i}, \pi^{\neg i}) % \pi^i \in \Pi(\pi^{\neg i *})
% \end{equation}
% for all agents $i$. 
Several works  \cite{ wang2021variational, so2022multimodal, kim2022maximum} show that the LNE policies satisfy the coupled Boltzmann distributions:
\begin{equation} \label{eqn:coupled_boltzmann}
    \pi^i(u|\mathbf x) = \exp\big(A^i(\mathbf x, u) \big) \quad \forall \pi^i
\end{equation}
where $A^i$ is the\textit{ advantage function}
\begin{equation} \label{eqn:advantage}
    A^i(\mathbf x, u) = \bar Q^i(\mathbf x, u) - V^i(\mathbf x),
\end{equation}
where the action-value function $\bar Q^i(\mathbf x_t, u_t^i)$ is defined as an expectation over the policies of all other agents $\neg i$, i.e.
\begin{equation} \label{eqn:best_response}
    \bar Q^i(\mathbf x, u) := \mathbb E^{\pi^{\neg i}} [Q^i(\mathbf x, \mathbf u)],
\end{equation}
and the value function $V^i(\mathbf x) := \log \int \exp \bar Q^i(\mathbf x, u) du$ serves as the \textit{log-partition function}. 
 % where $\mathbf u_t$ is the joint action available at time $t$. 
Eqn.~\eqref{eqn:best_response} reflects the \textit{best response} of agent $i$ to the expected actions of all other agents $\neg i$. As a consequence, the policy of agent $i$ given in Eqns.~\eqref{eqn:coupled_boltzmann} and \eqref{eqn:best_response} is a function only of the payoffs of the actions of agent $i$.

\begin{figure*}[t!]
    \centering
    \includegraphics[width=0.95\textwidth]{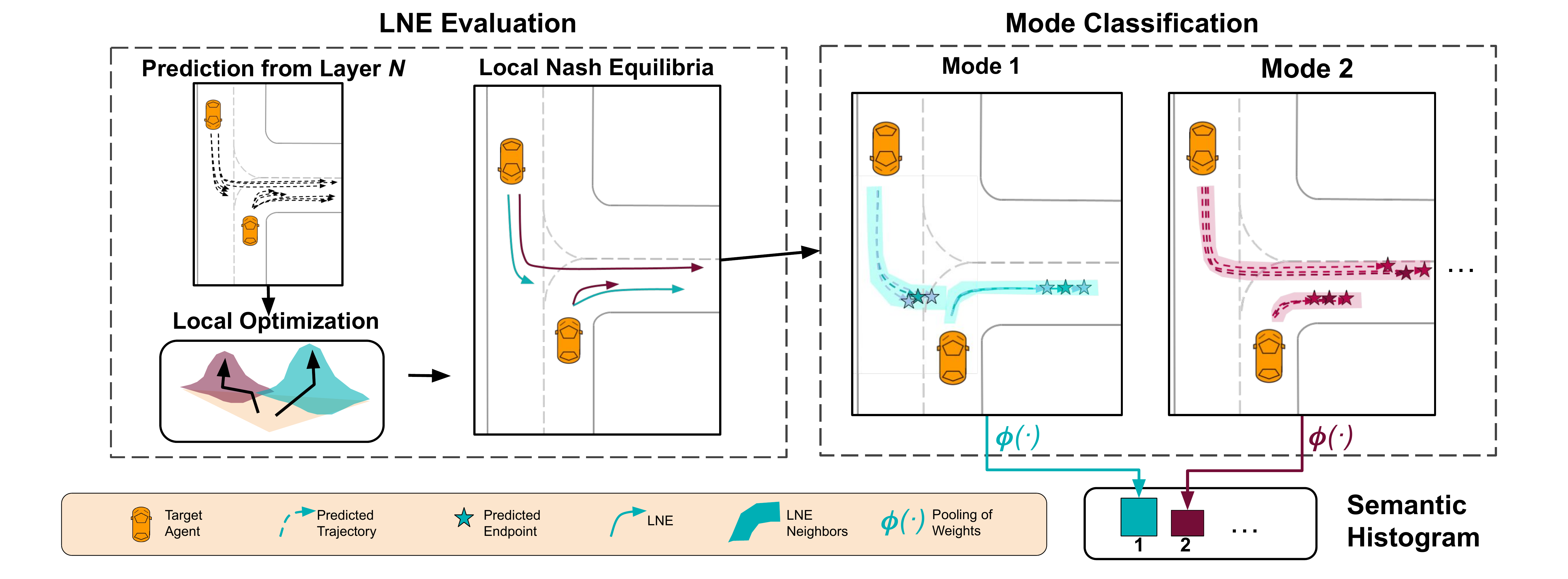}
    \caption{Game-theoretic analysis module. The prediction from the final transformer decoder layer is given as input to a local optimization algorithm \cite{cheng1995mean} to find the LNE. The closest LNE to each prediction are labeled, and the prediction weight belonging to each LNE are summed and normalized to create a histogram. \vspace{-0.2in}}
    \label{fig:coverage_optimization}
\end{figure*}

%
% In the maximum entropy framework, the value function %\eqref{eqn:value_function} 
%  for agent $i$ is defined as
% \begin{equation} \label{eqn:value_function}
%     V^i(\mathbf x_t) := \log \int \exp \bar Q^i(\mathbf x_t, u) du 
% \end{equation}
% and serves as the log-partition function in the MaxEnt setting \cite{mehr2021maximum}.
%\cite{haarnoja2017reinforcement}.  
% Each agent's policy satisfies the Markov property \cite{bacsar1998dynamic}
% % \begin{assumption} \label{assumption:markov}
% % (Markov Property) Each policy $\pi^i$ is a function of the joint state $\mathbf x_t$ conditionally independent of the joint states $(x_1, ...., x_{t-1})$ for all agents $i$ and time $t$. 
% % \end{assumption}
% % The partition function normalizes the policy distribution and is written as 
% % \begin{equation} \label{eq:partition}
% %     Z_t^i = \int \exp(\bar Q^i(\mathbf x_t, u_t^i)) d u_t^i 
% % \end{equation}
% and is estimated using the Laplace approximation \cite{so2022multimodal, mehr2021maximum} or importance sampling \cite{finn2016guided}. 
Next, we detail a learning procedure for Eqn.~\eqref{eqn:best_response}.
%, with careful attention to the partition function given in eq.~\eqref{eq:partition}.

% \vspace{-0.1in}

%\cite{ivanovic2021mats}, with other options also possible \cite{salzmann2020trajectron++}). 

% By the Markovian assumption, actions $u_t\sim \pi^i(\cdot|\mathbf x_t)$ are conditionally independent. 

% We adapt the sampling based IRL objective \cite{finn2016guided,phan2022driving}.
Given a dataset $\mathcal D$ of state-action histories of each agent (i.e., \textit{trajectories}), we regress a log-likelihood of a single agent's action, conditioned on the multi-agent history, as given by Eqn.~\eqref{eqn:coupled_boltzmann}. %While some works \cite{mehr2021maximum, so} explore methods for computing the coupled Boltzmann distributions, we 
%instead leverage the optimality condition of LNE as given by Assumption  \ref{assumption:LNE}.
%\begin{assumption} \label{assumption:LNE}
 %Assume that all agents in dataset $\mathcal D$ follow LNE.
%\end{assumption}
%Leveraging Assumption \ref{assumption:LNE}, 
We assume all agents in the data follow LNE and compute the best response in Eqn.~\eqref{eqn:best_response} so as to avoid the inner expectation in Eqn.~\eqref{eqn:coupled_boltzmann}. We maximize the log-probability of individual state-action pairs, which is equivalent to maximizing the advantage function:
\begin{equation} \label{eqn:item_loss}
   \log \pi^i(u|\mathbf x) = A^i(\mathbf x, u).
\end{equation}
Using the item loss \eqref{eqn:item_loss}, we seek to learn the multi-agent MaxEnt policy that maximizes the likelihood of the data:
\begin{equation} \label{eqn:dataset_irl_loss}
    \mathcal L_{IRL}(\mathcal D) = -\mathbb E^{\boldsymbol{\tau} \sim \mathcal D}\Bigg[\frac{1}{TI}\sum_{t=1}^T \sum_{i=1}^I \log \pi^i(u_t^i|\mathbf x_t) \Bigg] .
\end{equation}

\subsection{Enumerating Local Nash Equilibria} \label{appraoch: Mode Opt}

In this section, we detail how representative LNE are distilled in Fig.~\ref{fig:predictor_training}(a-c). By iteratively evaluating Eqn.~\eqref{eqn:coupled_boltzmann}, the log-likelihood of a joint trajectory $\boldsymbol{\tau} := (\tau^1, ..., \tau^I) = ((\mathbf x_1, \mathbf u_1), (\mathbf x_2, \mathbf u_2), ..., (\mathbf x_T, \mathbf u_T))$ is given as
\begin{equation} \label{eqn:log_likelihood}
   \log p( \boldsymbol{\tau}) = \sum_{t=1}^T \sum_{i=1}^I \log \pi^i(u_{t}^i|\mathbf x_{t}).
\end{equation}
under deterministic dynamics. 
% We show in Lemma 1 (See Appendix) that the product distribution \eqref{eqn:log_likelihood} is another Boltzmann distribution with a partition function equal to $1$. 
Under the LNE policies given by Eqn.~\eqref{eqn:coupled_boltzmann}, each mode of $p(\boldsymbol{\tau})$ (locally) maximizes the \textit{cumulative advantage} 
\begin{equation}\label{eqn:cumulative_advantage}
    A(\boldsymbol{\tau}):= \sum_{t=1}^T  \sum_{i=1}^I A^i(\mathbf x_t, u_t),
\end{equation}
i.e. the sum of the utilities of each agent along a joint trajectory. In the next sections, we detail how the representative LNE and their weights are computed using the Mean Shift algorithm (LNE Evaluation), how candidate trajectories are assigned to the ``nearest'' LNE (Mode classification ), and finally how the classified modes are summarized in a histogram according to their weights (semantic histogram). Fig.~\ref{fig:coverage_optimization} depicts this process.

\paragraph{LNE Evaluation} We employ the Mean Shift \cite{comaniciu2002mean,cheng1995mean} algorithm to explore the game-theoretic modes of the posterior distribution over trajectories, in terms of their sum of utility along the trajectory. The LNE evaluation module takes as input a set $\mathcal S$ of $K$ multi-agent trajectories emitted from the predictor. We evaluate the maximum-entropy likelihood of multi-agent trajectories, and iteratively seek distinct modes. To explore candidate trajectories, we utilize the dense outputs of the first $N$ transformer decoder layers, consisting of $K$ samples. Each output $k$ is scored using \eqref{eqn:cumulative_advantage} to compute $A(\boldsymbol{\tau}_k)$. The set of sampled scores $\{A(\boldsymbol{\tau}_k)\}_{k=1}^K$ is given as input to the Mean Shift optimizer and evaluated using their Boltzmann weights to compute the following kernel:
\begin{equation} \label{eqn: Mean Shift_kernel}
    p(\boldsymbol{\tau}, \boldsymbol{\tau}^\prime) = \exp (A(\boldsymbol{\tau}) - A(\boldsymbol{\tau}^\prime)) \mathbbm{1}\{d(\boldsymbol{\tau}, \boldsymbol{\tau}^\prime) < b\}).
\end{equation}
Intuitively, the first term in Eqn.~\eqref{eqn: Mean Shift_kernel} promotes modes that have a higher advantage (evaluated over a full trajectory) than other nearby modes. The second term in \eqref{eqn: Mean Shift_kernel} nullifies trajectories that are far away in norm, $d$ is the average (over agents) Euclidean distance between trajectory endpoints, and $b$ is a tunable bandwidth.

\paragraph{Mode Classification} The output of the Mean Shift algorithm gives a set of $M \leq K$ locally optimal multi-agent trajectories and  their weights $\{A(\boldsymbol{\tau}_m)\}_{m=1}^M$. From this set, we compute a set of labels $\boldsymbol{L}$ that assigns each prediction from the final decoder layer to the nearest LNE, using the nearest-neighbor algorithm. Each element $\boldsymbol{L}_m$ gives a one-hot embedding in $\{0, 1\}^M$. 

\paragraph{Semantic Histogram} The representative modes are used to compute a differentiable histogram. Given the LNE labels $\boldsymbol{L}$, and mode weights $\{A(\boldsymbol{\tau_m})\}_{m=1}^M$, the density corresponding to each LNE is computed as 
\begin{equation} \label{eqn: mode_hist}
    {q}^*_m(\boldsymbol{\tau}; \rho) = S\Bigg( \rho \cdot \phi(A(\boldsymbol{\tau}_m^1),..., A(\boldsymbol{\tau}_m^{N_m}))  \Bigg)
\end{equation}
where $N_m$ is the number of trajectories with label $m$, $\{\boldsymbol{\tau}_m^k\}_{k=1}^{N_m}$ are the classified trajectories closest to mode $m$, $\phi(\cdot)$ denotes max-pooling, $S(\cdot)$ denotes the softmax operation, and $\rho$ is a shape parameter. We refer to this histogram as the \textit{Boltzmann weights} of the representative LNE. %
%Henceforth, we use max-pooling as the pooling operation $\phi(\cdot)$. 
%Henceforth, we let $\phi(\cdot)$ represent max-pooling.
We show later in this section how Eqn.~\eqref{eqn: mode_hist} is used in the computation of a training-time auxiliary loss to promote sample diversity, wherein higher-entropy semantic histograms  indicate uniform coverage of each semantic mode found by Mean Shift, and hence sample diversity. 

 % We use scaled Gaussian as the Mean Shift kernel and 

% as target density. We use the drawn perturbation set, along with its sampling probabilities to perform the Mean Shift iterations similar to importance sampling \cite{finn2016guided}.  We model each mode as a Gaussian, the maximum number of modes $M$ and their mean and variance, with $M$ chosen in advance. In this work, we set $M=10$ such that the mode set may be larger sample set. If the total number of modes is less than $M$, we take $M$ to be that number instead. 
% Note that eqn. $\eqref{eqn: Mean Shift_kernel}$ is maximized when $S(\boldsymbol{\tau})$ is large. 

% , mode centers ${\boldsymbol{\tau}_m^*}_{m=1}^M$, LNE assignments for each point, and mode basins of attraction $\Sigma_m := \frac{1}{N_{mode}}\sum_{k=1}^K (\boldsymbol{\tau}-\boldsymbol{\tau}_m^*)^T(\boldsymbol{\tau}-\boldsymbol{\tau}_m^*)$, defined over each mode cluster of size $N_{mode}$. 

% An abridged version of  Mean Shift is presented in algorithm \ref{alg: Mean Shift}. 

% \begin{algorithm}
% \caption{Modified  Mean Shift  }\label{alg: Mean Shift}
% \begin{algorithmic}
% \INPUT{Set of trajectories $T$}
% \For{$n=1$ to \texttt{max-iter} or \texttt{converged}}
% \For{each point $\boldsymbol{\tau} \in T$}
%     \State Evaluate $w_\boldsymbol{\tau} \leftarrow p(\boldsymbol{\tau}) K(\boldsymbol{\tau}, \boldsymbol{\tau}_k) / q(\boldsymbol{\tau}) \quad \forall \boldsymbol{\tau}_k$
%     \State Update $\boldsymbol{\tau} \leftarrow \frac{\sum_{k=1}^Kw_k\boldsymbol{\tau}_k}{\sum_{k=1}^Kw_k}$
% \EndFor
% \EndFor
% \end{algorithmic}
% \end{algorithm}
% \vspace{-0.1in}

\subsection{Model Details} \label{Appraoch: model details}

\paragraph{Input Representation} Let $N$ be the number of agents in the scene for the scene and $I$ be the number of \textit{target} agents, for whom we make a prediction. For each focal agent $i \in [I]$, the transformer encodes the state and map information in the coordinate frame aligned with the heading angle of agent $i$. The state history is input as $X^i \in \mathbb R^{N \times T_0 \times C_{k}}$, where $T_0$ is the past horizon and $C_{k}$ is the coordinate dimension of the state histories that includes kinematic information, including position, velocity, acceleration, yaw, and yaw rate in agent-local coordinates. The map is input as $M^i \in \mathbb R^{N_s \times N_p \times C_m}$, where $N_s$ is the number of map segments, $N_p$ is the number of points per segment, and $C_m$ is the coordinate dimension of the map that includes the position, normal vector, and tangent vector of each map segment. Each agent's map input originally consists of $N_{m}$ points, which we then arrange into $N_s$ segments of at most $N_p$ points consisting of its nearest neighbors. We zero-pad segments of less than $N_p$ points.

\paragraph{Encoder and Decoder Structure} Agent and map features are computed, in the local coordinates of each target agent $i\in[I]$  and in parallel. As shown in Fig.~\ref{fig:predictor_training}, the agent and map encoders takes as input the observed agent features and map segments about the scene over a fixed past horizon $T_0$ in agent local coordinates. The encoder and decoder structure are identical to \cite{shi2022motion}.

%Our map encoder follows the form of \cite{2020vectornet}, iteratively applying local attention and max pooling within each map segment and global attention between segments. The outputs of the trajectory and map encoders are passed to an $L$-layer transformer encoder for global cross attention between state histories and map segment embeddings. Embedding are obtained in parallel for each target agent.  %Temporal edges are represented via edge modules, and nodes and self-edges are represented via an LSTM. We encode node and edge features via an MLP over normalized coordinates. 
%A single round of message passing between nodes is performed to propagate information over the network. 

% \paragraph{Decoder}

% Our decoder stacks $L$ transformer decoder layers for iterative mode refinement and one final transformer decoder layer for the final game-aware prediction. Each transformer decoder layer takes as input the $K$ static queries and cross-attends the static queries with trajectory and map segment embeddings. We use independent MLP decoders to compute ojoint predictions and the sample weights associated with each prediction. For each target agent, the output trajectories have shape $[S, T, C]$, $S$ is the sample size, $T$ is the number of future time steps, and $C=7$ is the GMM coordinate dimension, where each time point contains a scalar position, velocity, and log-variance in the $x$- and $y$-axis, and the mode confidence. 

%\vspace{-0.1in}

\paragraph{Sample Selection} 

In trajectory prediction, the number of trajectory samples is limited, typically to $S=6$. To increase the intrinsic diversity of samples we explore several post processing techniques: FPS, NMS, and a novel technique that suppresses non-LNE. FPS \cite{gonzalez1985clustering} iteratively maximizes the $l_2$-distance between samples. The NMS procedure \cite{zhao2021tnt} takes as input a set of weighted samples, and iteratively constructs a set of representative samples by taking the highest-weighted samples within a ball of a pre-defined radius. 

To encourage sampling of diverse interactive behaviors in the online setting, we introduce a new sampling method,\textit{ Non-Equilibrium Suppression} (NES), which functions similarly to NMS but uses the LNE assignments $\boldsymbol{L}$ defined in section III.B to suppress samples with the same mode assignment until all LNE have been sampled. 
%The procedure for NES is detailed in the appendix. 

% An additional approach for diversifying samples can be via an $L2$ loss term matching samples to the nearest neighbor from a set of FPS-samples from the predictor. We denote this loss by FPS Loss.

%While LSTM \cite{chai2020multipath} and Transformer \cite{Ngiam2021SceneTransformer, Nayakanti2022Wayformer} architectures offer better FDE/ADE, we choose an MLP decoder due to its (i)  fast convergence, and (ii) efficient integration into a game theoretic trajectory evaluation framework. In line with other trajectory prediction literature, we emit $K=6$ samples. See \cite{shi2022motion} for more detail on the decoder implementation. In expeirments, we set $N=6$ and $L=64$. \todo{check}. 

% \paragraph{IRL Utility Network}

% The IRL utility network (Fig.~\ref{fig:predictor_training}(b)) uses an encoder identical to the prediction module and takes as input the weighted trajectories, the weighted predictions and the current acceleration $u_t^i$, for each agent $i$ at time $t$. Actions are passed through an attention encoder and finally through an MLP to obtain a scalar utility. 

% \vspace{-0.1in}
\subsection{Loss Definitions}  \label{Approach: Loss Definitions}

We train NashFormer by jointly optimizing prediction accuracy, prediction classification, and mode coverage. 

\paragraph{Prediction Loss} 

In line with other multimodal prediction literature that use weighted trajectory sets \cite{Ngiam2021SceneTransformer, Varadarajan2021MultipathPlusPlus,shi2022motion}, we chose for each agent a multi-agent Gaussian mixture model with $K$ components that represents the marginal likelihood of an observed point $o_t^i = (x_t^i, y_t^i)$:
\begin{equation} \label{eqn:gmm_likelihood}
    P^i(o_t^i) = \sum_{k=1}^K p_{k, t}^i  f_{k, t}^i (x_t^i - \mu_{x, t}^i, y_t^i - \mu_{y, t}^i)
\end{equation}
where $f_{k, t}^i(\cdot, \cdot)$ is the component (Guassian) likelihood of the $i$th agent's observation $o_t^i$, and $p_{k, t}^i$ is the occurrence probability of the $i$th agent's motion mode.

The accuracy loss is given as the mean cumulative log-likelihood of the closest component $h^i$ for each agent $i$ over the time horizon $T$, i.e. 
\begin{equation} \label{eqn:mon_loss}
    \mathcal L_{acc} = - \frac{1}{I} \sum_{i=1}^I \sum_{t=1}^T \Big (\log p_{k, t}^i  + \log f_{k, t}^i (x_t^i - \mu_{x, t}^i, y_t^i - \mu_{y, t}^i) \Big).
\end{equation}
We choose the index where the average ground truth distance of each agent's endpoints to the motion mode endpoints are minimized. We also use an $l_1$ velocity loss and a dense prediction loss as in \cite{shi2022motion}.

\paragraph{Classification Loss}

% Rather than optimizing eqn. \eqref{eqn:dataset_irl_loss} directly to learn the \textit{action}-level policies, we first learn the distribution over the marginal \textit{trajectories} of each agent, equivalent to the summation over time in eqn. \eqref{eqn:dataset_irl_loss} for fixed agent index $i$ as a cross-entropy classification loss, defined as
As in \cite{Ngiam2021SceneTransformer, Varadarajan2021MultipathPlusPlus,shi2022motion}, we learn a weight distribution $w$ that describes the marginal likelihood of each motion mode. The marginal likelihoods are learned using a cross entropy loss
% We show in Lemmas 1 and 2 (See Appendix) that IRL loss \eqref{eqn:dataset_irl_loss} is approximately equivalent to a sample-based cross-entropy loss. 
\begin{equation} \label{eqn:classification loss}
    \mathcal L_{class} = - \sum_{k=1}^K \mathbbm 1\{k = \hat k\} \log w_k,
\end{equation}
which penalizes trajectories that are weighted highly but classified as different motion modes. The marginal log-probabilities are then sorted and summed to get the joint trajectory likelihood, i.e., the full summation in Eqn.~\eqref{eqn:dataset_irl_loss}. 

% \paragraph{Prediction Loss} In line with other multimodal prediction literature that use weighted trajectory sets, we include in the data term both an Minimum-over-N (MoN) loss \todo{replace with GMM}. \cite{gupta2018social,thiede2019analyzing}
% \begin{equation} \label{eqn:mon_loss}
%     \mathcal L_{acc} = \min_{\boldsymbol{\tau}^{(k)} \in \mathcal S} ||\boldsymbol{\tau}^{(k)} - \hat \boldsymbol{\tau}||_2,
% \end{equation}
% and the classification loss \cite{kawasaki2021multimodal},
% \begin{equation} \label{eqn:classification loss}
%     \mathcal L_{class} = - \sum_{k=1}^K \mathbbm 1\{k = \hat k\} \log w_k.
% \end{equation}
% $\hat \boldsymbol{\tau}$ is the ground truth trajectory, and the index $\hat k$ is given as the closest trajectory to the ground truth as measured by the $L2$ distance. $\mathbbm 1(\cdot)$ is an indicator function.

% \paragraph{Classification Loss}We additionally apply a classification loss \cite{cui2019multimodal} to encourage accurate weighting of samples.
% The classification loss is defined as the cross-entropy of the discrete distribution over samples:
% where $\mathbbm 1(\cdot)$ denotes the indicator function.

\paragraph{Coverage Loss} To ensure maximal representation of the learned LNE in the final samples, we propose a novel coverage loss $\mathcal L_{cov}$, given as the KL-divergence between discrete semantic histogram \eqref{eqn: mode_hist} and the histogram obtained from the emitted trajectory set $\mathcal S$, described next. Similar to Eqn.~\eqref{eqn: mode_hist}, we approximate a discrete distribution $q_m$ over the mode support $[1,...,M]$, given as the softmax operation
\begin{equation} \label{eqn:softmax_predicted}
        q_m(\mathcal S) = S\Bigg(\sum_{k=1}^K A(\boldsymbol{\tau}_k) \mathbbm{1}\{\boldsymbol{L}_m = k\}\Bigg)
\end{equation}
For each mode $m$, Eqn.~\eqref{eqn:softmax_predicted} sums the contribution of each sample to the empirical mode likelihood. 

The predicted mode distribution in Eqn.~\eqref{eqn:softmax_predicted} is contrasted with the mode distribution according to the game-theoretic (Boltzmann) model, which is approximated discretely as Eqn.~\eqref{eqn: mode_hist}.
where $\rho\rightarrow 0$ results in a uniform distribution for $q^*$ over the top $M$ game-theoretic modes). 
The predictor's coverage loss is thus defined as the KL-divergence between $q$ and $q^*$, i.e. 
\begin{equation} \label{eqn:coverage_loss}
    \mathcal L_{cov} = D_{KL}[ \; q \;|| \; q^* \;].
\end{equation}

\paragraph{Final Loss}
The final loss is computed as
\begin{equation} \label{eqn: overall_loss}
    \mathcal L = \alpha \mathcal L_{acc} + \beta \mathcal L_{class} + \gamma \mathcal L_{cov}. %+ \lambda_2 \mathcal L_{regression},
\end{equation}
Therefore, for $\gamma=0$, our loss is equivalent to standard losses for learning a set of weighted trajectories, such as those used in \cite{shi2022motion, Varadarajan2021MultipathPlusPlus, Ngiam2021SceneTransformer}. However, for $\gamma>0$, our novel coverage loss $\mathcal L_{cov}$ also allows the user to prioritize coverage of LNE.  

\begin{table}
    \centering
    \setlength\tabcolsep{2pt}
    \begin{tabular}{c|c|cccc}
    \hline
        \rowcolor{Gainsboro!60} & Method & minADE $\downarrow$ & minFDE  $\downarrow$ & Miss Rate $\downarrow$ & mAP $\uparrow$\\
        \hline
        \multirow{8}{*}{Test} & LSTM Baseline \cite{ettinger2021waymo} & 1.096 & 5.028 & 0.775 & 0.052 \\
        & HeatIRm4 \cite{mo2021heterogeneous}& 1.420 & 3.260 & 0.722 & 0.084 \\
        & AIR$^2$ \cite{wu2021air} & 1.317 & 2.714 & 0.623 & 0.096 \\
        & DenseTNT \cite{gu2021densetnt} & 1.142 & 2.490 & 0.535 & 0.165 \\
        & M2I \cite{sun2022m2i} & 1.351 & 2.833 & 0.554 & 0.124 \\
        & Scene Trans. \cite{Ngiam2021SceneTransformer} & 0.977 & 2.189 & 0.494 & 0.119 \\
        & MTR \cite{shi2022motion} & 0.918 & 2.063 & 0.441 & 0.204 \\ 
        & GameFormer (J) \cite{huang2023gameformer} & 0.916 &  1.937 & 0.441 & 0.137 \\
        \hline
        \multirow{3}{*}{Val}  & MTR \cite{shi2022motion} & 0.913 & 2.054 & 0.437 & 0.199 \\ 
        & \textit{NashFormer-Base (Ours)} & 0.961 &2.198 & 0.508 & 0.192\\
        & \textit{NashFormer-GT (Ours)} & 0.978 & 2.238 & 0.495 & 0.182 \\
        & \textit{NashFormer-NES (Ours)} & 0.944 & 2.146 & 0.483 & 0.193 \\
    \end{tabular}
    \caption{\underline{Joint} metrics comparing game-theory-agnostic and game-theory model at 8$s$. Accuracy metrics are in meters. \vspace{-.2in}}
    \setlength\tabcolsep{6pt}
    \label{tab:baselinesv2}
\end{table}

\begin{figure*}[t]
    \centering
    \begin{tabular}{ccc}
        \subfigure[\textbf{Yield.} Agents 1 and 2 exhibit a multi-modal yield interaction either $1$ or $2$ may turn towards the lower street, causing the other to break. Other interactions are also possible, including agent $1$ going straight, affecting agent $2$'s timing. ]{\includegraphics[width=0.31\linewidth]{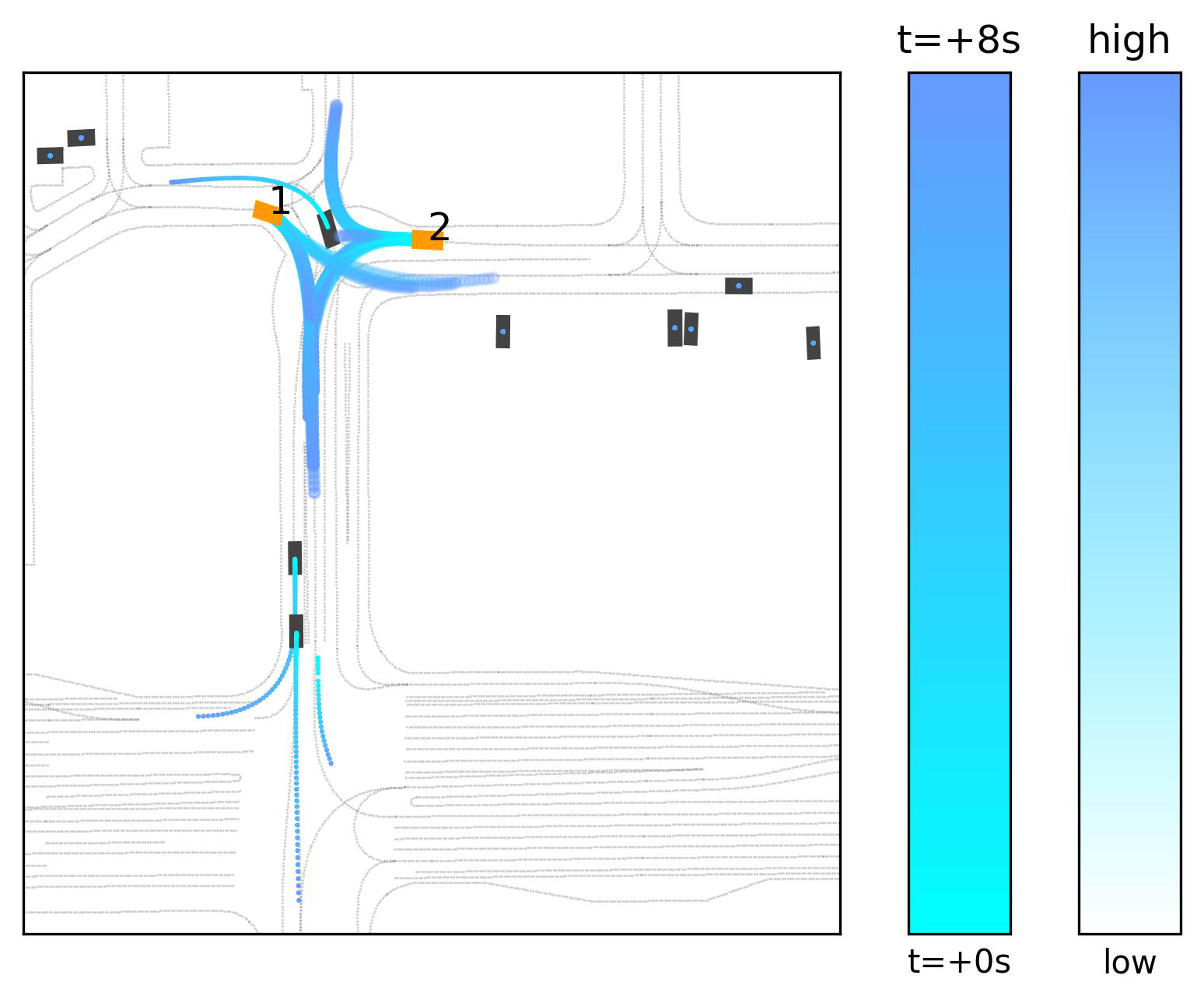}} & 
        \subfigure[\textbf{Follow.} Agent $2$ turns left onto the lower street but must consider agent $1$'s intent: if agent $1$ continues straight, agent $2$ can turn at a high right of speed, but if agent $1$ also turns, agent $2$ must follow at a lower velocity. ]{\includegraphics[width=0.31\linewidth]{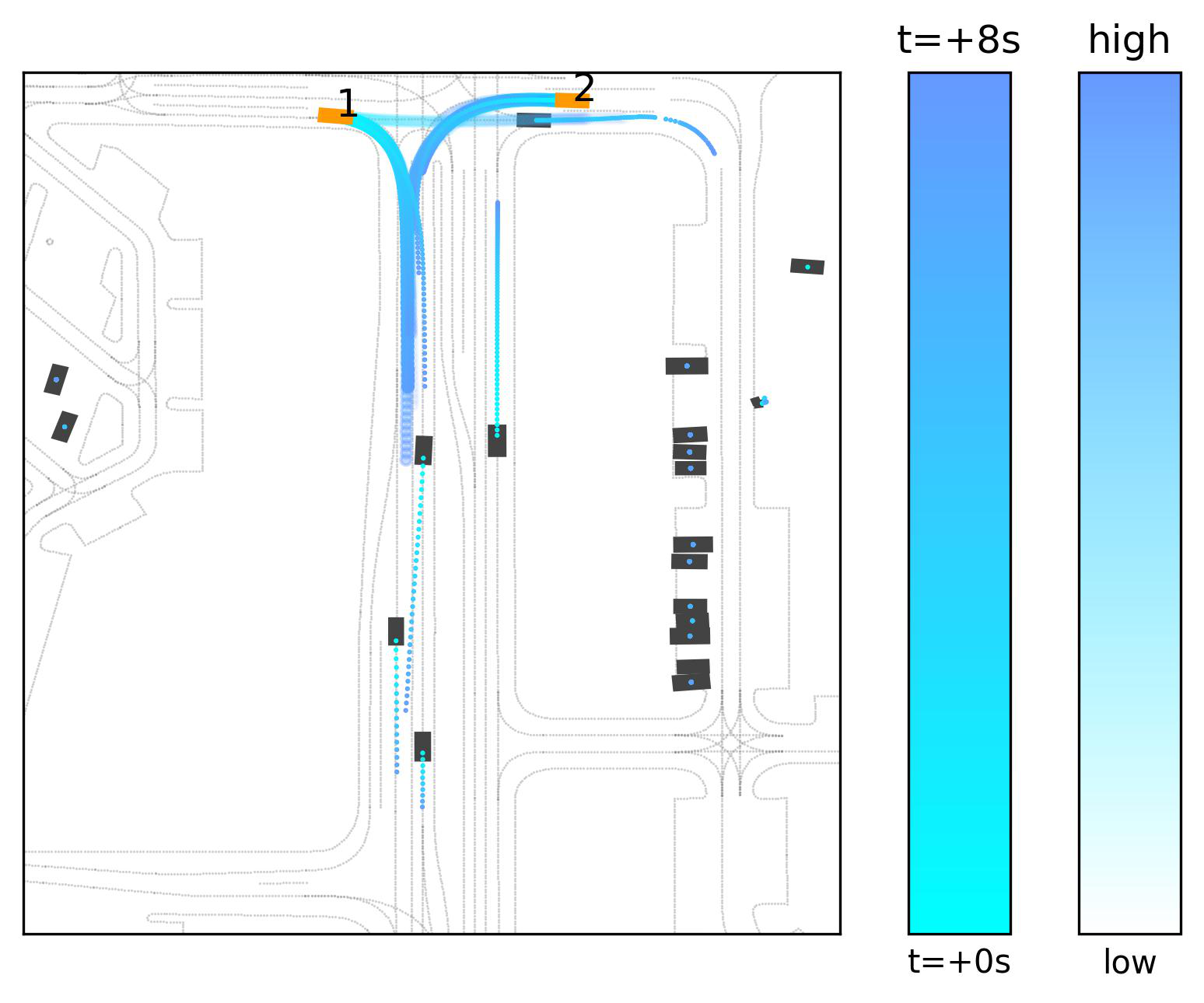}} &
        \subfigure[\textbf{Low-TTC.} Three plausible risk-varying choices where agent $2$ turns onto a busy street: turning out immediately in close proximity to other vehicles, waiting briefly for the middle lane, or waiting patiently for traffic to clear. ]{\includegraphics[width=0.31\linewidth]{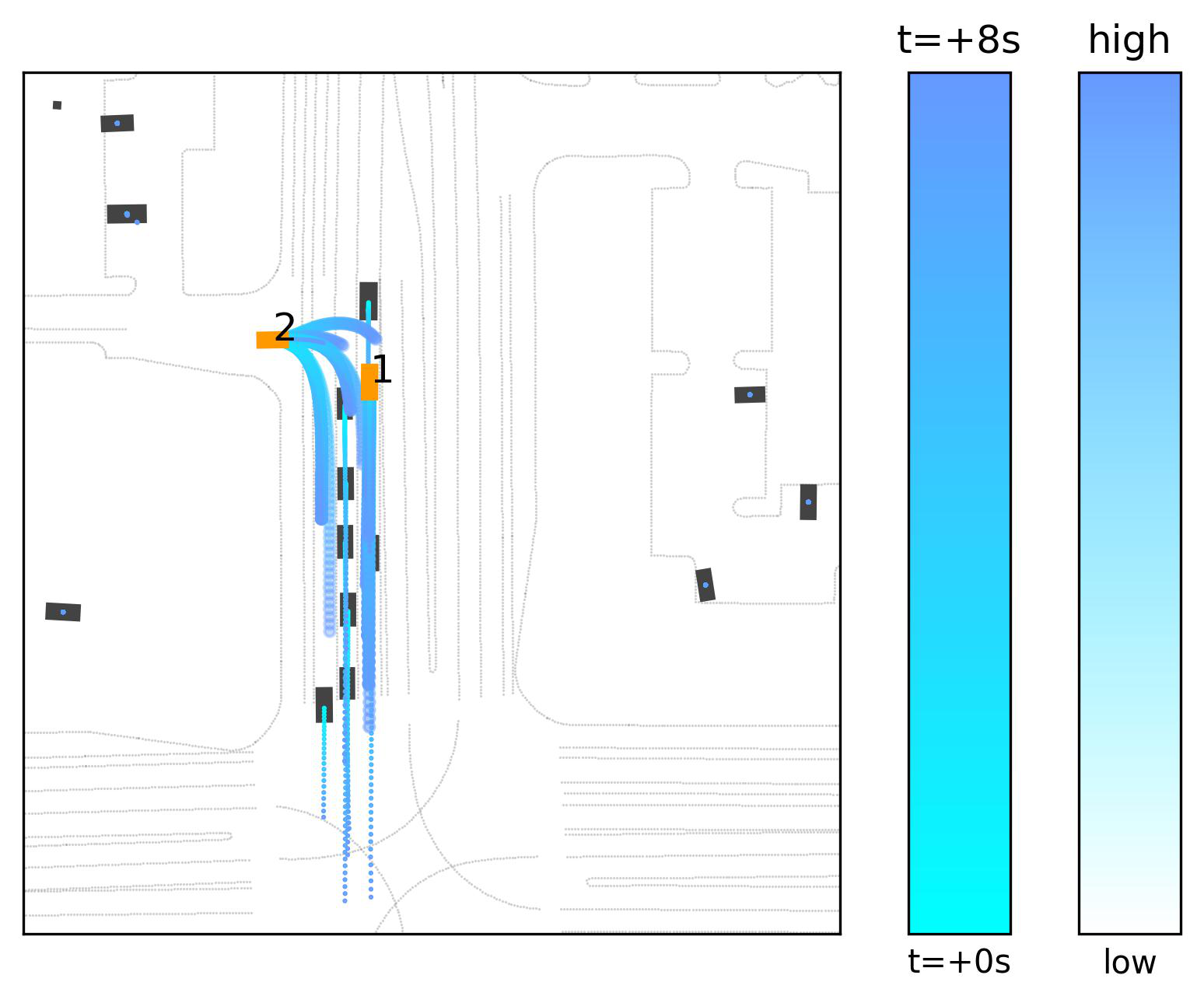}} 
    \end{tabular}
    \caption{Bird's eye view trajectories with interactions between agents $1$ and $2$. NashFormer balances predictive weighting of multiple-equilibria game-theoretic interactions, resulting in better coverage of various outcomes. \vspace{-.2in}}
    \label{fig:bev_baseline}
\end{figure*}

\section{Experiments}

We now introduce the dataset and experiments demonstrating
the effectiveness of our approach compared to several baselines.

\subsection{Experiment Design}

% \vspace{-0.1in}
\paragraph{Dataset Description}
We evaluate our model on the interactive split of the large-scale Waymo Open Motion Dataset (WOMD) \cite{ettinger2021waymo}, which has a diverse set of real-world traffic scenes. 
% Morover, the interactive split contains the largest number of interactive driving scenes to data.
The diversity of interaction, and long prediction horizon make it especially challenging for prediction. The interactive prediction task requires the model to predict the joint future of two target agents, with up to 128 total agents per scene. The dataset provides one second of trajectory history per agent, and the prediction horizon is eight seconds. There are $487k$ training scenes and $44$k validation scenes for the interactive challenge. As metrics, we compute scene-averaged MoN average and final displacement errors (minADE, minFDE), miss rate, and mean average precsion (mAP), defined in \cite{ettinger2021waymo}. 

\paragraph{Model Implementation} For each focal agent of the target agents to predict, the state histories of all other agents and the map are first normalized to the coordinate frame of the focal agent. The transformer structure is identical to that of \cite{shi2022motion}. For the overall loss \eqref{eqn: overall_loss}, we use $\alpha=1$, $\beta=1$, and $\gamma=10$. We use $b=10$ as the Mean Shift bandwidth parameter. 

%The target and context embeddings are obtained using an MLP with input dimension $31$, which includes position, velocity, acceleration, and one-hot position encoding. The map segment embedding is obtained using an MLP with input dimension $50$, which includes the position of each map \textit{vector} (i.e. two sequential points), the tangent and normal vectors, and the one-hot point class index. Individual segments are max-pooled, then cross-attended similar to \cite{gao2020vectornet}. The hidden dimension for the history embeddings is $256$ and for the map embedding is $64$. In the transformer encoder, we stack $6$ transformer encoder layers to cross-attend between target, context, and history embeddings, similar to other works \cite{kumar2020interaction, Ngiam2021SceneTransformer, gu2021densetnt, shi2022motion}. For decoding, we stack $N=3$ transformer decoder layers. We pre-process all target, context, and map embeddings using an MLP with output dimension $512$ for each history embedding and motion query and $256$ for the map embeddings. Similar to \cite{shi2022motion}, we use an MLP with hidden dimension $1280$ and output dimension $512$ to fuse decoder features. Each transformer-decoder layer takes as input the $L=100$ motion queries and cross attends the embeddings form the encoder to obtain $L=100$ samples. For NMS, we set the distance threshold to $2.5m$ as in \cite{shi2022motion}. 

\paragraph{Coverage evaluation} We evaluate the coverage of semantic interactions, or \textit{semantic diversity}, by the entropy of the weights assigned to a set of (post-hoc) interaction labels $q$ \cite{huang2020diversitygan} on the validation set,
\begin{equation} \label{eqn:semantic_coverage}
    \mathcal H_{\mathcal S} = -\sum_{m=1}^M q_m \log q_m,
\end{equation}
where $M:=|\mathcal S|$ is the total number of semantic interaction modes, $q_m := z_m^{-1}\sum_{k=1}^K \mathbbm 1 [\tau_k \in \mathcal S_m]w_k$ is the sum of the predictor weights assigned to interaction mode $m$, $z_m$ is a normalizing constant, and $\mathcal S_m$ is the set of trajectories that satisfy a label $m$, such as a specific interaction category, such as ``$A$ yielding to $B$''. In this work, we compute several semantic diversity metrics for the final samples selected by NMS: $\mathcal H_{\textnormal{Util.}}$ for the cumulative advantage, $\mathcal H_{\textnormal{Yield}}$ for yield scenarios, $\mathcal H_{\textnormal{Follow}}$ for follow scenarios, $\mathcal H_{\textnormal{TTC}}$ for low-proximity maneuvers. \textit{Importantly, we do not provide the interaction labels during training or during inference.}  In experiments we report the dataset-averaged $H_{[\cdot]}$ as defined in \eqref{eqn:semantic_coverage} to gauge coverage.

%We test our model in two stages. First, we train a baseline model for joint trajectory prediction using only the accuracy \eqref{eqn:mon_loss} and classification \eqref{eqn:classification loss} losses. In parallel, the game theoretic utility model is trained using \eqref{eqn:dataset_irl_loss}. Once both models are converged, we compute the coverage loss \eqref{eqn:coverage_loss}. 

\subsection{WOMD Interactive Prediction Task}

\noindent
\textbf{Quantitative Results.} Table \ref{tab:baselinesv2} details the performance of our model on the Waymo Interactive Dataset \cite{ettinger2021waymo} against several recent \cite{mo2021heterogeneous, shi2022motion, huang2023gameformer} baseline models that do not use an additional coverage loss \eqref{eqn:coverage_loss} as in our approach. We present scores for three models: Nashformer-Base, which does not use the additional coverage loss \eqref{eqn:coverage_loss} and employs NMS sampling at a standard threshold of $2.5$ meters; NashFormer-GT, which uses the game-theoretic coverage loss \eqref{eqn:coverage_loss} along with NMS (also at $2.5$ meters) to increase coverage of multiple equilibria; and NashFormer-NES, which uses NES sampling in addition to the coverage loss to prioritizing sampling of multiple equilibria.  Against the validation set, our base model performs similarly to the successful MTR \cite{shi2022motion} model in minADE, minFDE, and mAP, with the NES model enjoying a slight boost in performance due to inference-time reasoning about game-theoretic equilibria. 

Both GT and NES variants of NashFormer enjoy a lower miss rate than the base model, which supports the hyporthesis that generic predictors may emit redundant modes, even with NMS sampling, as presented in Fig.~\ref{fig:teaser}.

\begin{table*}[h!]
    \setlength\tabcolsep{2.5pt}
    \centering
    \begin{tabular}{c|ccc|cccccccc}
        \hline
        \rowcolor{Gainsboro!60} & GT Loss & Method & Thresh.  &  KL$\downarrow$ & minADE $\downarrow$  & $N_{\textnormal{modes}}$ $\uparrow$  & $\mathcal H_{\textnormal{Util.}}$ $\uparrow$ & $\mathcal H_{\textnormal{Yield}}$ $\uparrow$  & $\mathcal H_{\textnormal{Follow}}$ $\uparrow$  & $\mathcal H_{\textnormal{TTC}}$ $\uparrow$  \\
        \hline 
        \multirow{4}{*}{\textit{Game-Agnostic}} & \xmark & FPS (top 24) & -- & 0.302 & 1.370 & 1.512 & 0.182 & 0.035 & 0.018 & 0.018 \\
        & \xmark & FPS (top 12) & -- & 0.302 & 1.168 & 1.529 & 0.573 & 0.090 & 0.053 & 0.055 \\
        & \xmark & NMS & 2.5 & 0.302 & 0.961 & 1.575 & 1.193 & 0.142 & 0.094 & 0.086 \\
        & \xmark & NMS & 10.0 & 0.301 & 1.166 & 1.546 & 0.700 & 0.142 & 0.073 & 0.072\\
        \hline
        \multirow{4}{*}{\textit{Game-Aware}} 
        & \cmark & NMS & 2.5  & 0.104 & 0.984 & 2.069 & 1.362 & \textbf{0.162} & 0.120 & 0.120 \\
        & \cmark& NMS & 10.0 &  0.104 & 1.203 & 2.018 & 1.015 & 0.146 & 0.116 & \textbf{0.129}\\
        & \xmark & NES  & 10.0 & 0.303 & \textbf{0.944} & 1.544& 1.228 & 0.142 & 0.096 & 0.086 \\
        & \cmark & NES  & 10.0 & \textbf{0.103}& 0.978 & \textbf{2.150} & \textbf{1.374} & \textbf{0.162} & \textbf{0.121} & \textbf{0.129} \\
    \end{tabular}
    \caption{Effect of game-aware loss $\mathcal L_{cov}$ and sampling strategies versus game-agnostic methods. We compare NashFormer with and without the coverage loss \eqref{eqn:coverage_loss} and additionally explore several coverage-inducing sampling strategies: FPS with the top $k$-weighted samples (here, $k$ is $12$ or $24$), NMS with both a small ($2.5$m) and large ($10$m) threshold, and NES, presented earlier. Game-theoretic analysis improves coverage of local equilibria for all metrics. \vspace{-0.2in}}
    \label{tab:ablation}
\end{table*}

\begin{figure}
    \centering
    \includegraphics[width=0.24\textwidth]{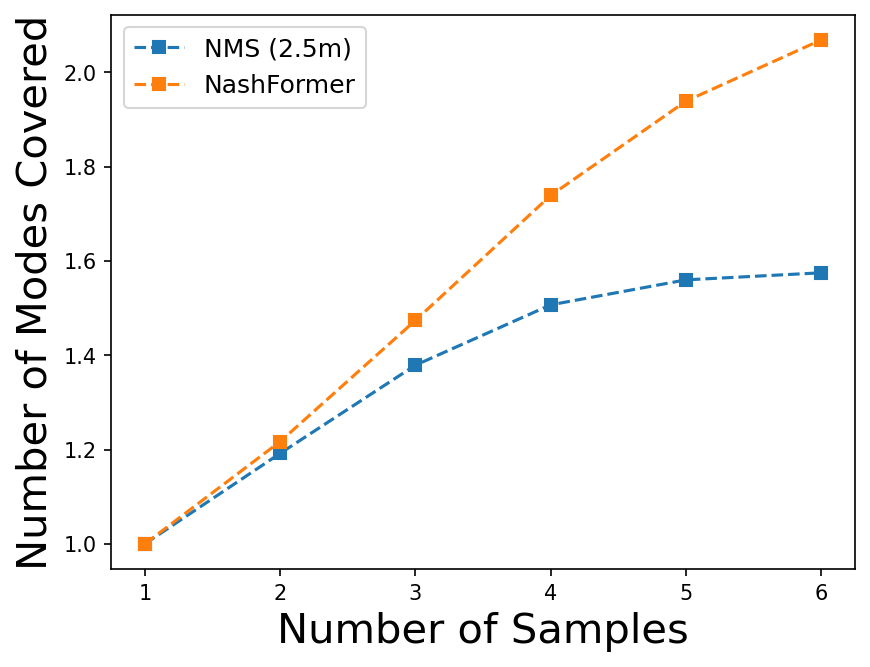}
    \includegraphics[width=0.24\textwidth]{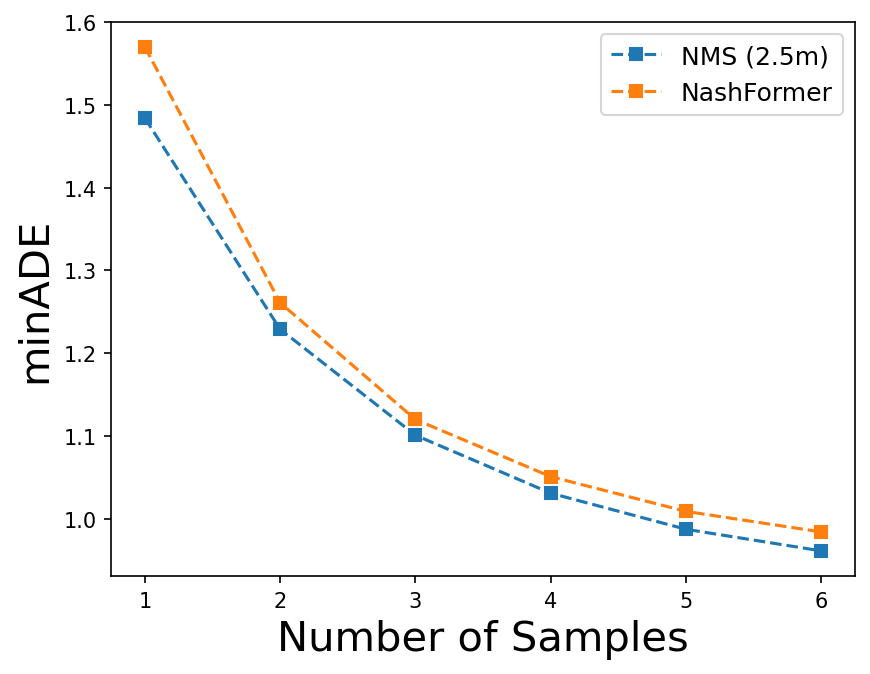}
    \caption{Mode coverage (left) and minADE (right) as the number of prediction samples increases for the full NashFormer model versus an off-the-shelf predictor with NMS sampling at a 2.5 meter threshold. NashFormer permits each sample to contribute to coverage, while the off-the-shelf predictor plateaus around the fourth sample. Moreover, NashFormer does not suffer significant loss of minADE accuracy as it's permitting better coverage. \vspace{0in}} %  \textit{without} a plateau.}
    \label{fig:cov_fde}
\end{figure}

\noindent
\textbf{Ablation Study.} Table \ref{tab:ablation} presents an ablation study on the effect of both the game-theoretic coverage loss \eqref{eqn:coverage_loss} and sampling strategy on the coverage of local equilibria for three coverage-inducing strategies: FPS, NMS, and NES. We present the raw KL score for \eqref{eqn:coverage_loss} in addition to the prediction minADE, number of prediction modes covered $N_{\textnormal{modes}}$, and post-hoc semantic diversity scores for four interaction categories: raw utility, yielding, following, and low-TTC (close-proximity driving). Our results show that generic sampling methods such as FPS and NMS do not automatically capture game-theoretic equilibria; only NashFormer, using knowledge of the global cost landscape, captures multiple distinct LNE. For NES, we present a game-aware variant of a generic predictor (second to last row) that only uses NES without a game-theoretic loss to capture a higher number equilirbia. We note that the lower number of modes covered $N_{\textnormal{modes}}$ indicates that the coverage loss \eqref{eqn:coverage_loss} may result in finding \textit{new equilibria}; nonetheless, semantic diversity scores are still improved using only the equilibria available to the base model. In Fig.~\ref{fig:cov_fde}, we show that NashFormer better captures LNE using fewer samples than a generic predictor using NMS.  

We also performed an ablation of the effect of the coverage loss coefficient $\gamma$ on the minADE and number of modes covered. The base NashFormer model using NMS at a $2.5$ thresholded covered $1.546$ modes with a minADE score of $0.98$ (see Table 2). By varying $\gamma\in[0, 100]$, the number of modes covered by NashFormer is in the range $[1.55,3.50]$ and sacrifices only up to in minADE, in the range $[0.98, 1.07]$. Compared to other approach such as FPS that naively increase pairwise distance between samples and sacrifice minADE, NashFormer prioritizes diverse \textit{equilibria} that are intrinsically local optima in the posterior trajectory distribution.

% \begin{table}
%     \centering

%     \begin{tabular}{l|cccc}
%     \hline
%         \rowcolor{Gainsboro!60} Bandwidth & minADE $\downarrow$ &\textbf{Miss Rate} $\downarrow$ & $N_{\textnormal{modes}}$ $\uparrow$ & $\mathcal H_{\textnormal{Util.}}$ $\uparrow$  \\
%         \hline 
%         1 & & & & \\
%         2.5 & & & & \\
%         5 & & & & \\
%         10 & & & & \\
%     \end{tabular}
%     \caption{Effect of Mean Shift local optimization bandwidth (in meters) on prediction accuracy and coverage.}
%     \label{tab:baselines}
% \end{table}

\begin{figure}
    \centering
    \includegraphics[width=0.24\textwidth]{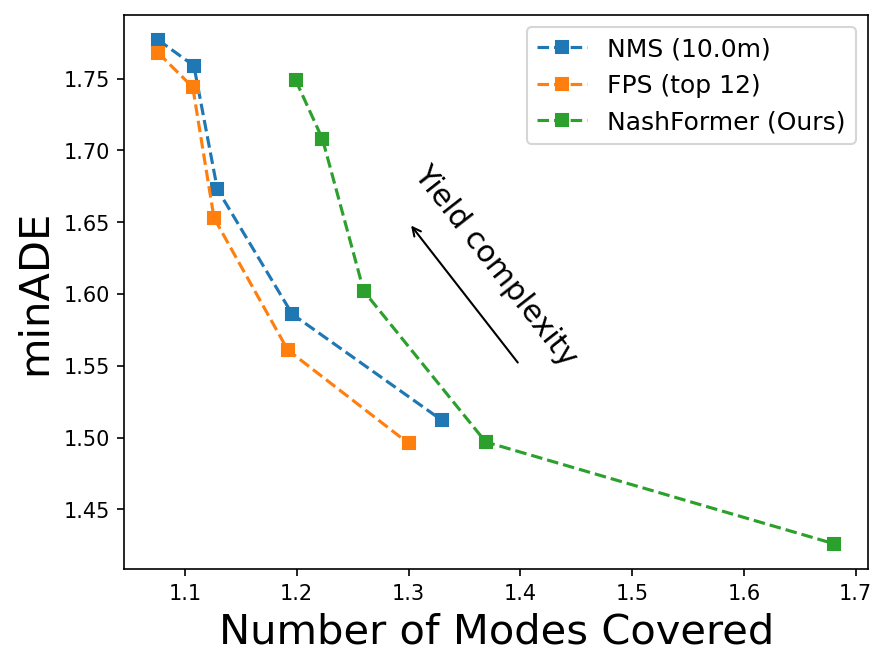}
    \caption{Coverage versus minADE (meters) for NashFormer versus baseline models. NashFormer achieves up to $33.6$ percent higher coverage of semantic modes while minADE is largely unaffected by the increased sample diversity. \vspace{-.2in}}
    \label{fig:splits}
\end{figure}

\noindent
\textbf{Qualitative Results} Our experimental results show that NashFormer achieves higher coverage of scenario-specific modalities versus a baseline without game-awareness. As demonstrated in Fig.~\ref{fig:bev_baseline}, NashFormer improves coverage of empirically rare scenarios on the basis of their existence as equilibria, providing insight into different possible outcomes. In Fig.~\ref{fig:bev_baseline}(a), multiple yield interactions are possible, including at various timings, depending on the choices of agents $1$ and $2$. In Fig.~\ref{fig:bev_baseline}(b), multiple follow interactions are possible between agents $1$ and $2$ depending on if agent $1$ chooses to go forward or turn right. In Fig.~\ref{fig:bev_baseline}(c), agent $2$ may choose various risky behavior depending on their perceived cumulative advantage; it is important for other road agents, especially agent $1$, to consider these choices or risk violating close-proximity safety protocols. While the above results hold for the WOMD interactive dataset, which has fixed interaction complexity, we show in the following section that NashFormer can better generalize to even more challenging interactive scenarios with \textit{increasing} interaction complexity.

\subsection{High-Complexity Yielding Scenarios}

We demonstrate the efficacy of our model on several high-complexity \textit{network} interactions that specifically involve yielding. These scenarios capture NashFormer's capability to cover distinct yielding outcomes when decision uncertainty is high. 

\noindent
\textbf{Filtering for Complex Network Interactions} To evaluate coverage of game-theoretic outcomes, we augment the WOMD training set with labels indicating high-level yield interactions. Specifically, we consider two agents $i$ and $j$ as nodes in a directed graph $G$ to satisfy $\texttt{yield(i,j)}$ interaction if agent $i$ yields to agent $j$. We evaluate \textit{network} interactions in which there occurs \textit{at least} $k$ yield interactions, for $0 \leq k \leq 4$. 

\noindent
\textbf{Results} Fig.~\ref{fig:splits} shows the trend in minADE as scenario complexity increases. NashFormer, which intrinsically samples diverse LNE, enjoys lower minADE versus other diverse sampling methods (FPS, NMS). We note that the number of modes covered \textit{decreases} for all model as the complexity of interaction increases; empirically, as the number of nodes in a network interaction grows, so does the basin of attraction of multi-agent equilbria. In other words, fewer stable multi-agent equilibria may exist when agents interact in a spatially compact group. Nonetheless, NashFormer covers a more diverse set of modes while scoring lower minADE versus the other diversity-promoting methods, showing that the other methods have poor prioritization of LNE.

% \begin{table}
%     \centering

%     \begin{tabular}{l|cccc}
%     \hline
%         \rowcolor{Gainsboro!60} Number of Yields & minADE $\downarrow$ &\textbf{Miss Rate} $\downarrow$ & $N_{\textnormal{modes}}$ $\uparrow$  & $\mathcal H_{\textnormal{Util.}}$ $\uparrow$  \\
%         \hline 
%         0 & & & & \\
%         1 & & & & \\
%         2 & & & & \\
%         3 & & & & \\
%     \end{tabular}
%     \caption{Effect of dataset minimum yield count (per scenario) on on prediction accuracy and coverage.}
%     \label{tab:network_yield_performance}
% \end{table}

\section{Conclusion}
In this paper we introduce \textit{NashFormer}, a prediction model that uses local Nash equilibria (LNE) to improve the coverage of prediction networks without presuming a specific maneuver taxonomy, and beyond what is afforded by metric trajectory coverage approaches. Our experiments reveal coverage of semantic interactions, lending insight into different modes of utility-driven decision-making. We believe our game-theoretic prediction framework opens several research avenues for more effective game-theoretic planning.

% \newpage

%\Todo{check style}
\bibliographystyle{IEEEtran}
\bibliography{refs} 

\onecolumn
\setlength\parindent{0pt}
\newpage
%\section*{Appendix}
%\include*{appendix/lemma1}
%\include*{appendix/lemma2}
%\include*{appendix/implementation_details}
%\include*{appendix/scenario_details}
%\include*{appendix/additional_qualitative}
%\include*{appendix/nomenclature}

\end{document}